\documentclass[letterpaper,10pt,journal,twoside]{IEEEtran}

\usepackage{graphicx} 
\usepackage{amsmath}
\usepackage{amssymb}
\usepackage{booktabs}
\usepackage{url}
\usepackage{comment}
\usepackage{hyperref}
\usepackage{cite}
\usepackage[table,xcdraw]{xcolor}
\definecolor{black}{rgb}{0, 0, 0}
\definecolor{red}{rgb}{0.9, 0, 0}
\definecolor{green}{rgb}{0, 0.6, 0}
\definecolor{blue}{rgb}{0, 0, 0.9}
\definecolor{grey}{rgb}{0.52, 0.52, 0.51}

\newcommand{\BLACK}[1]{\textcolor{black}{#1}}

\newcommand{\rev}[1]{\BLACK{}{#1}}

\title{FRA-NBV: A Fast and Reflectivity-Aware Next-Best-View Strategy}

\author{G.~F.~Preziosa$^{1}$, E.~Setti$^{1}$, M.~Faroni$^{1}$, A.~M.~Zanchettin$^{1}$, and P.~Rocco$^{1}$%
\thanks{$^{1}$The authors are with Politecnico di Milano, 20133 Milano, Italy ({\footnotesize e-mail: giuseppefabio.preziosa@polimi.it}).}%
\thanks{Digital Object Identifier (DOI): see top of this page.}}

\begin{document}

\maketitle

\begin{abstract}
Autonomous 3D reconstruction with depth sensors is strongly affected by reflective surfaces, which cause missing or unreliable measurements and reduce the effectiveness of conventional Next-Best-View (NBV) strategies. This limitation is particularly critical in industrial applications involving reflective components and low-cost, low-resolution depth sensing, where robustness to sensing failures is essential. This paper proposes a Fast Reflectivity-Aware Next-Best-View (FRA-NBV) strategy that explicitly addresses reflection-induced depth loss without relying on prior object models or assumptions on material reflectance, making it suitable for a wide range of industrial configurations. Reflective regions are identified from the spatial distribution of missing depth measurements and localized in three-dimensional space using an online ellipsoid-based representation of the object estimate. A recovery strategy then selects additional poses that modify the sensor’s angle of incidence to improve the likelihood of reconstructing the affected regions.
Experiments on objects with different geometric and reflective complexity demonstrate that the approach significantly improves reconstruction coverage under realistic industrial conditions.


\end{abstract}


\section{INTRODUCTION}

\IEEEPARstart{A}{utonomous} 3D reconstruction is crucial in industrial environments characterized by flexibility and frequent reconfiguration. In such scenarios, robots should handle objects with varying geometry and pose, making manual reconfiguration impractical. 
To address these limitations, object reconstruction is typically achieved through multiple acquisitions from different sensor poses, allowing the sensing process to adapt to the observed object. This naturally leads to the Next-Best-View (NBV) problem, which focuses on selecting informative sensor poses for efficient and incremental 3D reconstruction \cite{review_NBV}.

In real industrial scenarios, not all objects are equally suited to depth-based measurements, as material properties and surface finishes—particularly those of metallic components commonly found in industrial environments—can severely degrade sensing quality. Specifically, depth sensors operate reliably on diffusely reflecting surfaces, whereas highly reflective and specular materials often degrade measurement quality by inducing image saturation or low signal-to-noise ratio, leading to missing or unreliable depth measurements \cite{reflection_1,reflection_3}. This limitation is further amplified in scalability-oriented solutions, where low-cost, low-resolution depth sensors are adopted to reduce system complexity and deployment costs. Although such sensors can be used for 3D reconstruction \cite{uf_low_res}, their larger noise and lower measurement reliability exacerbate the effects of challenging surface properties, leading to more frequent reconstruction failures and reduced effectiveness of traditional NBV strategies.

This paper deals with the need to identify surface regions that systematically degrade depth measurements and to actively guide the reconstruction process toward their recovery. Existing solutions address this challenge either by relying on external coating procedures to improve sensing reliability \cite{coating_reflective}, or by exploiting multi-view NBV strategies that compensate for missing data through sensor pose selection \cite{reflection_1}. However, these approaches typically assume prior knowledge of object geometry or material properties, restricting their applicability to known scenarios.
In contrast, we introduce a novel Fast Reflectivity-Aware Next-Best-View (FRA-NBV) strategy, with the following main contributions:
\begin{itemize}
    \item A novel method for identifying reflective regions by exploiting the spatial distribution of missing depth measurements, without relying on any prior knowledge of object geometry, material properties, or reflection models. The proposed approach leverages an online ellipsoid-based representation of the object estimate to localize missing measurements in three-dimensional space and infer their spatial extent. This compact representation enables reliable reasoning even with sparse and low-resolution depth sensors.


    \item A recovery strategy that selects additional sensor poses aimed at increasing the probability of successfully acquiring the identified reflective regions. Since no assumptions are made on object geometry or material characteristics, recovery is driven by varying the sensor pose to induce different angles of incidence thereby modifying the interaction between the sensor and the surface and improving the likelihood of reconstructing previously missing geometry.
\end{itemize}
Experimental results on objects with progressively increasing geometric and reflective complexity demonstrate the effectiveness of the proposed framework. In highly reflective scenarios, FRA-NBV improves reconstruction coverage by up to 31\% over the projection-based method and up to 56\% over the entropy-based method, while remaining comparable on non-reflective objects.
A video overview of the method, along with the reconstruction of both a reflective and a non-reflective object, is available at \url{https://merlinlaboratory.github.io/FRA-NBV/}.

\section{RELATED WORK}

The Next-Best-View problem addresses the selection of informative sensor poses for object reconstruction and has been studied since early works \cite{nbv_1993, nbv_199, nbv_2003}.
According to Lee et al.~\cite{lee}, prior NBV research can be decomposed into spatial representation, view search space, and view determination, with research objectives increasingly focusing on computational efficiency while meeting industrial constraints such as low-cost, low-resolution sensing and the presence of highly reflective objects.

\subsubsection{Spatial Representation}
It determines how acquired data are organized and updated. Among the most widely adopted approaches, volumetric representations discretize the environment into voxel grids, where each voxel encodes the occupancy probability.
This formulation allows voxels to be classified as free, occupied, or unknown, enabling efficient updates, visibility reasoning, and the identification of frontier voxels for NBV planning \cite{paper_0,frontier_2,frontier_3}.
Alternatively, surface-based approaches offer higher geometric accuracy at the cost of increased complexity, either relying on point clouds commonly used in learning-based NBV methods \cite{learning_1,learning_2}, or on reconstructed meshes \cite{mesh_1,mesh_2}.
More recently, implicit neural representations \cite{INR_1,INR_2} and 3D Gaussian Splatting (3DGS) \cite{gs_1,gs_2,gs_3} have emerged as alternatives to classical 3D representations, enabling continuous scene modeling.
In practice, voxel-based volumetric methods remain a common choice for NBV, providing a favorable trade-off between simplicity, efficiency, and integration. Following this formulation, in this work we adopt a voxel-based volumetric representation as the primary spatial model, continuously updated by integrating multi-view depth acquisitions.

\subsubsection{View determination}
It estimates the potential information gain associated with a candidate sensor pose and is closely tied to the adopted spatial representation.
For a volumetric representation, this process is commonly formulated through utility functions that quantify the volumetric information gain \cite{uf_learning}.
Entropy-based formulations are widely adopted in this context \cite{uf_entropy_1,uf_entropy_2}, often augmented with surface related terms \cite{uf_entropy_plus_area}.
In addition, quality-aware utility functions have also been proposed to address low-resolution and high-noise sensing conditions \cite{uf_quality_1}, while \cite{uf_low_res} specifically targets low-resolution sensors by prioritizing less-sampled regions and penalizing redundant views.
However, information gain is predominantly evaluated via ray-casting, whose high computational cost has motivated a range of acceleration strategies, including efficient voxel traversal \cite{uf_fast_ray_1}, hierarchical multi-resolution ray-casting \cite{uf_area_factor}, and ray tracing with local refinement \cite{uf_fast_ray_2}.
Rather than accelerating ray-based evaluation, learning-based approaches aim to substitute ray-casting: NBV-Net \cite{uf_learning} with a voxel-based 3D CNN, while PC-NBV \cite{uf_learning_2} learns the utility of candidate views directly from the partial point cloud. Despite promising results, these methods often require high computational resources and training on high-resolution datasets, limiting their applicability to low-resolution sensors. More recently, several works have improved computational efficiency by moving away from voxel-based ray-casting altogether and adopting continuous scene representations.
In this context, methods based on 3DGS \cite{gs_2,gs_3} enable fast view evaluation by leveraging image-space projections, effectively bypassing volumetric ray traversal.
However, these approaches rely primarily on RGB information, limiting their applicability in depth-only sensing scenarios.
In contrast, other methods retain the advantages of voxel-based spatial representations while introducing a secondary, higher-level representation specifically for view determination.
In this setting, the voxel-based scene is approximated using an ellipsoid-based representation obtained via Gaussian Mixture Model (GMM) clustering \cite{paper_0}.
This hybrid strategy accelerates view determination while preserving compatibility with pure depth sensors.
Moreover, it enables the integration of sparse measurements typical of low-resolution sensing by filling density gaps and alleviating the redundancy inherent to voxel-based ray-casting. These considerations motivate the adoption of hybrid voxel–projection formulations in this work. Nevertheless, regardless of the adopted representation or view determination strategy, low-cost depth sensing remains highly sensitive to measurement noise, particularly in the presence of reflective surfaces, which continue to pose a fundamental challenge.

\subsubsection{Reflective Surfaces}
Reflectivity poses two main challenges for NBV-based reconstruction: identifying regions affected by reflection-induced depth loss and selecting sensor poses that can reliably restore the missing geometry.
Yang et al. \cite{reflection_1} use a Phong-based reflection model to detect reflective regions and guide view selection. The method relies on CAD-derived surface hypotheses, which restrict applicability to known objects.
Ouyang et al. \cite{reflection_2} detect reflective areas from RGB images and restore missing depth with a trained neural network. The approach depends on a dedicated dataset and shows limited generalization.
Lee et al. \cite{reflection_3} infer surface reflectivity from infrared signal intensity in ToF cameras. However, reliable IR measurements are difficult to obtain with low-cost sensors.
Finally, \cite{gs_3} explores multimodal sensing by combining vision and tactile feedback to recover regions that are difficult to reconstruct visually, such as reflective surfaces.
However, tactile sensing is employed only as a post-processing step, after the visual reconstruction is largely completed.
Consequently, large reflective regions can still dominate visual view selection, resulting in redundant sensor poses and reduced reconstruction efficiency.

\rev{In summary, FRA-NBV extends PB-NBV~\cite{paper_0} by introducing two novel components: reflective-region identification and reflective-region restoration. Unlike prior reflective-aware approaches, the proposed method does not require CAD priors, material models, RGB information, or task-specific training data.}

\section{Preliminaries}
\label{sec:preliminaries}
The Next-Best-View problem is commonly formulated as a sequential decision process in which, at each iteration $t$, a finite set of candidate sensor poses
$\mathcal{X}_t = \{x_1, x_2, \dots, x_N\} \subset SE(3)$
is generated and evaluated through a utility function $U(x)$. The pose maximizing such utility is then selected to guide the next acquisition.

In this work, we build upon the ellipsoid-based NBV framework originally introduced in~\cite{paper_0}, which replaces computationally expensive ray-casting with a projection-based evaluation of clustered volumetric regions. 
We briefly summarize the framework adopted in~\cite{paper_0} in terms of spatial representation (Sec.~\ref{sec:spatial_representation}), object approximation (Sec.~\ref{sec:Ellipsoid Representation}) and fast view sampling evaluation (Sec.~\ref{sec:view_projection}). 

\subsection{Spatial Representation and Object Identification}
\label{sec:spatial_representation}

To represent data from multi-view depth observations, a voxel-based structure is employed, in which each voxel \rev{with a resolution $s_v$} encodes the occupancy probability.
As measurements are integrated over time, voxels are classified as free $\mathcal{V}_{\mathrm{fr}}$, occupied $\mathcal{V}_o$, or uncertain $\mathcal{V}_u$. Among uncertain voxels, \emph{frontier voxels} $\mathcal{V}_f \subset \mathcal{V}_u$ are defined as those lying at the boundary between free and occupied space and represent the most informative regions for exploration.
At iteration $t$, the object model $\mathcal{O}_t$ is extracted from the observed data by clustering the occupied voxels $\mathcal{V}_o$, yielding a compact spatial representation of the object. The spatial extent of $\mathcal{O}_t$ is described by an Oriented Bounding Box (OBB) $\mathcal{B}_t$, which provides a geometric support for view generation.

\subsection{Ellipsoid Representation}
\label{sec:Ellipsoid Representation}

A complementary description of the object estimate $\mathcal{O}_t$ is obtained by approximating its spatial structure with a limited number of ellipsoidal geometric primitives.
Starting from the voxel-based representation, occupied $\mathcal{V}_o$ and frontier voxels $\mathcal{V}_f$ are clustered independently using GMM, yielding the cluster sets $\mathcal{C}_o = \{ \mathcal{C}_o^{k} \}_{k=1}^{K_o}$ and $\mathcal{C}_f = \{ \mathcal{C}_f^{k} \}_{k=1}^{K_f}$. 
The numbers of mixture components $K_o$ and $K_f$ are selected automatically using the Bayesian Information Criterion (BIC), which enforces a trade-off between model complexity and data fidelity.
This enables the representation to adapt to the evolving geometry of $\mathcal{O}_t$ while preventing overfitting due to an excessive number of small ellipsoids, as well as underfitting caused by too few, overly coarse components that would poorly approximate the object shape.
Each cluster in $\mathcal{C}_o$ and $\mathcal{C}_f$ is subsequently approximated by a minimum-volume enclosing ellipsoid, yielding the ellipsoid sets $\mathcal{E}_o = \{ \mathcal{E}_o^{k} \}_{k=1}^{K_o}$ for the occupied regions and $\mathcal{E}_f = \{ \mathcal{E}_f^{k} \}_{k=1}^{K_f}$ for the frontier regions.

\begin{figure*}[!t]
\begin{center}
\includegraphics[width=17.5cm]{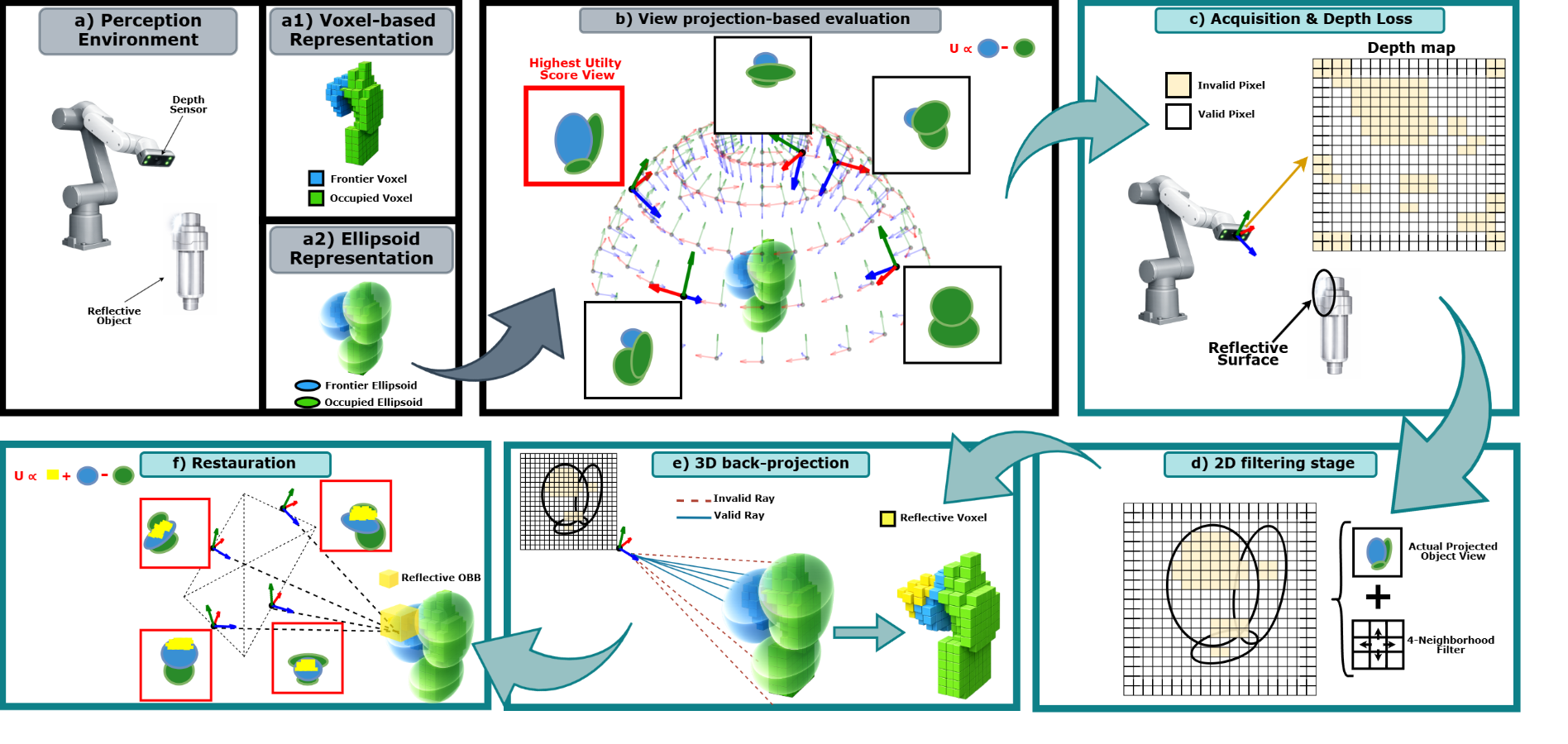}    
\vspace{-10pt}
\caption{Overview of the FRA-NBV pipeline. A voxel-based model and its ellipsoidal approximation are maintained and updated for efficient view evaluation (a–b). After acquisition (c), reflective regions are detected from invalid depth measurements and localized in 3D (d–e). These regions guide a recovery strategy that samples tilted sensor poses to improve reconstruction (f). \rev{The pipeline components inherited from PB-NBV \cite{paper_0} are shown in black, while the novel reflectivity-aware modules introduced in this work are highlighted in turquoise.}}
\label{fig:complete_pipeline}
\end{center}
\vspace{-5mm}
\end{figure*} 

\subsection{View Sampling and Projection-Based Evaluation}
\label{sec:view_projection}


Candidate sensor poses are generated using a sampling-based strategy. \rev{Based on the current OBB estimate $\mathcal{B}_t$, $N_{\text{pos}}$ poses are sampled within a hemispherical region centered on it, with radius $\gamma$ plus the half-diagonal of $\mathcal{B}_t$, and evaluated through a projection-based strategy that operates on the ellipsoidal representation introduced in Sec.~\ref{sec:Ellipsoid Representation}.}
For a candidate pose characterized by the camera projection matrix $\mathbf{P} = \mathbf{K}[\mathbf{R}\mid\mathbf{t}]$, the projection of an ellipsoid onto the image plane is obtained through:
\begin{equation}
\boldsymbol{\Phi}^{-1} = \mathbf{P}\mathbf{Q}^{-1}\mathbf{P}^\top,
\label{eq:ellipsoid_projection}
\end{equation}
where $\boldsymbol{\Phi}$ denotes the conic corresponding to the projected ellipsoid in the image plane, and $\mathbf{Q}$ is the quadric matrix encoding the ellipsoid’s spatial extent and orientation.
The utility of a candidate sensor pose is computed by aggregating the weighted projected areas of frontier ellipsoids and penalizing those corresponding to occupied regions,
\begin{equation}
U(x) =
\sum_{\mathcal{E}_f} w_j\,A_j
-
\sum_{\mathcal{E}_o} w_j\,A_j,
\label{eq:view_utility}
\end{equation}
where $A_j$ denotes the projected area of the $j$-th ellipsoid and $w_j$ is a weighting term that accounts for occlusions by favoring ellipsoids whose centers are closer to the camera. Specifically, ellipsoids are ordered according to their distance from the camera, and higher weights are assigned to nearer ellipsoids, while farther ones receive progressively lower weights. The next best view is selected as the candidate maximizing the utility $U(x)$. 


\section{Problem Statement}
\label{sec:prob_statement}

In the classical Next-Best-View formulation, the objective is to maximize the reconstructed surface given a limited number of sensor poses, or equivalently to minimize the number of views required to achieve a desired reconstruction level. When dealing with highly reflective objects, reflective regions may cause missing depth measurements. 
If such missing data are treated as standard unexplored geometry, classical NBV strategies—based solely on maximizing object coverage (cf. Eq.~(2))—tend to select redundant sensor poses that fail to resolve these regions, leading to inefficient reconstruction.
\rev{Prior work on reflective surface reconstruction typically exploits prior knowledge, such as CAD models, material properties, or explicit reflectance assumptions, to optimize the NBV process and identify reliable acquisition poses in advance~\cite{reflection_1}. In contrast, we consider the reconstruction of completely unknown objects, for which no prior information is available to analytically predict the most reliable sensor poses. Among the factors that remain directly controllable during acquisition, the incidence angle is therefore the most relevant one. Consequently, acquisition reliability must be modeled as an implicit, view-dependent property.}
In particular, given a sensor pose $x$ and the set of depth measurements performed from the same sensor pose:
\(
\mathcal{D}_t(x) = \{ \mathcal{D}_t^{(i)}(x) \}_{i=1}^{N_k},
\)
we denote by
\begin{equation}
F(S_k, \mathcal{D}_t(x)) \in \{0,1\},
\end{equation}
the function that indicates whether a surface element $S_k \subset O_t$ is responsible for missing depth measurements under the selected sensor pose. 
Our goal is to approximate and guide view selection toward the set of surfaces $S_r^t(x)$, defined as:
\begin{equation}
S_r^t(x) = \left\{ S_k \subset O_t \mid F(S_k, \mathcal{D}_t(x)) = 1 \right\},
\end{equation}
rather than treating missing measurements as standard unexplored geometry.



\section{Method}
\label{sec:method}
The complete pipeline is illustrated in Fig.~\ref{fig:complete_pipeline}. 
At each acquisition step, the scene is modeled through two complementary representations: a voxel-based occupancy structure (Fig.~\ref{fig:complete_pipeline}a1) and its ellipsoidal approximation (Fig.~\ref{fig:complete_pipeline}a2). 
These models are incrementally updated and enable the projection-based evaluation of candidate sensor poses (Fig.~\ref{fig:complete_pipeline}b), through the utility function defined in Eq.~\eqref{eq:view_utility}.
As shown in Fig.~\ref{fig:complete_pipeline}c, acquiring depth measurements in the presence of reflective surfaces, often result in unreliable or missing data.
For this reason, we extend~\cite{paper_0} with two additional stages: we first detect reflective regions (Sec.~\ref{sec:reflective_identification}, Fig.~\ref{fig:complete_pipeline}c--e), and then use them to guide view planning (Sec.~\ref{sec:reflective_restoration}, Fig.~\ref{fig:complete_pipeline}f).


\subsection{Identification of Reflective Regions}
\label{sec:reflective_identification}

To identify reflective surface regions $S_r^t(\hat{x})$ associated with a fixed sensor pose $x=\hat{x}$, we analyze missing depth values. \rev{In our setting, invalid depth measurements mainly originate from three causes: points lying outside the working range of the sensor, sensor noise, and reflective surfaces. Therefore, the objective of the proposed detector is not to label every missing depth measurement as reflective, but rather to identify persistent and spatially coherent invalid patterns that are compatible with reflection-induced sensing failures under the assumptions of our reconstruction scenario.}

\rev{To achieve this goal, a sequence of filtering stages is applied. The first stage aims at reducing the influence of isolated sensor noise by identifying pixels that are repeatedly invalid across the $N_k$ depth acquisitions performed from the same pose.}
In particular, each depth acquisition $\mathcal{D}_t^{(i)}$ is represented as a depth map defined over the image domain $\Omega$ of size $W \times H$,
\begin{equation}
\mathcal{D}_t^{(i)} = \{ z_{u,v}^{(i)} \in \mathbb{R} \mid (u,v) \in \Omega \}
\end{equation}
where $z_{u,v}^{(i)}$ denotes the depth measurement associated with pixel location $(u,v)$ in the $i$-th acquisition. Missing or invalid depth measurements are represented by zero-valued entries in the depth map. Based on this, a pixel-wise invalidity ratio is computed across the $N_k$ acquisitions as:
\begin{equation}
\eta(u,v) = \frac{1}{N_k} \sum_{i=1}^{N_k} \mathbb{I}\!\left({z}_{u,v}^{(i)} = 0\right),
\label{eq:invalid_ratio}
\end{equation}
where $\mathbb{I}(\cdot)$ is the indicator function. Pixels whose invalidity ratio exceeds 0.5 are classified as frequently invalid, yielding the confidence map $\mathcal{I}$:
\begin{equation}
\mathcal{I} = \{ (u,v) \in \Omega \mid \eta(u,v) \ge 0.5 \}.
\label{eq:invalid_set}
\end{equation}

\rev{The confidence map $\mathcal{I}$ may still contain invalid pixels caused by factors other than reflectivity, such as points outside the sensor working range. Therefore, a second 2D filtering stage exploits the current estimate of the object extent.}
To this end, as shown in Fig.~\ref{fig:complete_pipeline}d, the ellipsoidal representation of the object introduced in Sec.~\ref{sec:Ellipsoid Representation} is projected onto the image plane of the current sensor pose $\hat{x}$ following the projection model described in Sec.~\ref{sec:view_projection}. The resulting binary image mask $\mathcal{I}_o \subset \Omega$, obtained from the combined projections of the ellipsoids $\mathcal{E}_o \cup \mathcal{E}_f$, approximates the object silhouette under the current camera pose. \rev{Pixels outside the projected silhouette are discarded by intersecting the confidence map $\mathcal{I}$ with $\mathcal{I}_o$.}
\rev{However, the ellipsoidal approximation may slightly overestimate the object extent, causing some boundary pixels to be incorrectly associated with the object silhouette. These invalid measurements are more likely due to points beyond the sensor working range than to reflective effects. To mitigate this issue, a third filtering stage is required.}
Let $\mathcal{N}_4(p)$ denote the 4-connected neighborhood of a pixel $p=(u,v)$; a pixel is retained as a candidate reflective pixel if:
\begin{equation}
\mathcal{I}_c :=
\left\{
p \in \mathcal{I} \cap \mathcal{I}_o
\;\middle|\;
\forall q \in \mathcal{N}_4(p),\; q \in \mathcal{I}_o
\right\}.
\end{equation}

\rev{At this stage, a final filtering stage exploits the ellipsoidal representation of the object to recover their three-dimensional location.}
For each pixel \(p=(u,v)\in\mathcal{I}_c\), let
\begin{equation}
\mathbf{r}(t) = \mathbf{c}_{\mathbf{c}} + t\,\mathbf{d}_w, \quad t > 0,
\label{eq:viewing_ray}
\end{equation}
be a viewing ray,
where $\mathbf{c}_{\mathbf{c}} \in \mathbb{R}^3$ denotes the ray origin, corresponding to the camera optical center. The ray direction $\mathbf{d}_w \in \mathbb{R}^3$ is derived from the image coordinates \((u,v)\) as follows:
\begin{equation}
\mathbf{d}_c =
\Bigg[
\dfrac{u - c_x}{f_x},
\dfrac{v - c_y}{f_y},
1
\Bigg]^T,
\qquad
\mathbf{d}_w = \mathbf{R}_{cw}\,\mathbf{d}_c .
\end{equation}
where \((f_x,f_y)\) and \((c_x,c_y)\) denote the focal lengths and the optical center (in pixels), respectively, and \(\mathbf{R}_{cw}\) is the rotation matrix that transforms vectors from the camera frame to the world frame. As depicted in Fig.~\ref{fig:complete_pipeline}e, the viewing ray $\mathbf{r}(t)$ is intersected with all ellipsoids $\mathcal{E}_o \cup \mathcal{E}_f$ representing the object. Among all valid intersections, the closest one to the camera is selected, representing a plausible estimate of the surface point that would have been observed in the absence of reflective effects. Pixels whose rays do not intersect any ellipsoid, or whose first intersection occurs with an ellipsoid associated with occupied regions, are discarded. In this way, only points lying on frontier ellipsoids are retained. \rev{This step rejects pixels that are classified as reflective in the image plane but correspond to surface regions that have already been reconstructed from previous sensor poses.} The remaining set of 3D points is grouped using the DBSCAN \cite{DBSCAN} algorithm, yielding a set of clusters $\mathcal{C}_r = \{ \mathcal{C}_r^{k} \}_{k=1}^{K_r}$ that correspond to spatially coherent reflective regions. Each cluster $\mathcal{C}_r^{k}$ is enclosed within an OBB $\mathcal{B}_r^{k}$, producing a set of three-dimensional regions $\mathcal{B}_r = \{ \mathcal{B}_r^{k} \}_{k=1}^{K_r}$ that provide a discrete, view-dependent estimate of the surface set $S_r^t(\hat{x})$.

\begin{figure*}[t]
\centering
\begin{minipage}[c]{0.73\textwidth}
\centering
\includegraphics[width=\linewidth]{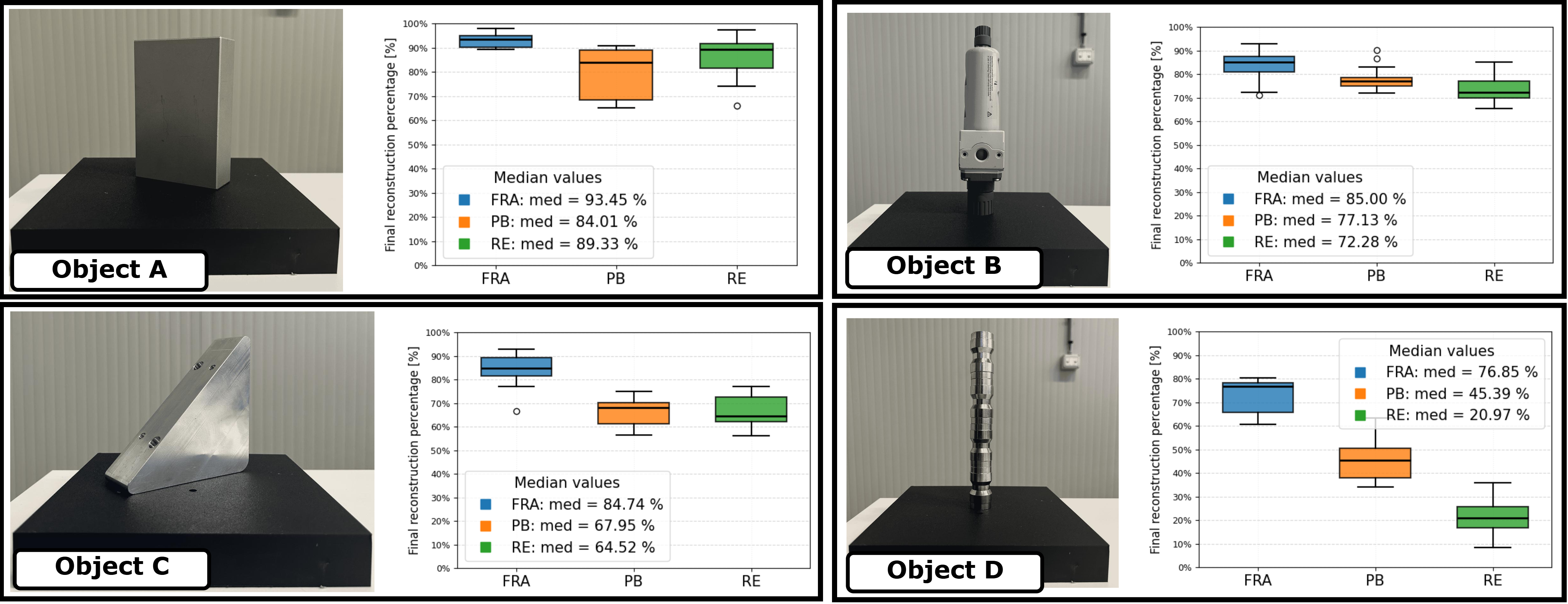}\\[0.5mm]
\textbf{(a)}
\end{minipage}\hfill
\begin{minipage}[c]{0.25\textwidth}
\centering
{\scriptsize
\setlength{\tabcolsep}{3.5pt}
\renewcommand{\arraystretch}{1.1}
\begin{tabular}{cc}
\toprule
\textbf{Parameter} & \textbf{Value} \\
\midrule
\(N_{\max}\) & 15 \\
\midrule
\(\tau_{\mathrm{stag}}\) & \(1\%\) (for 5 acq.) \\
\midrule
\(s_v\) & \(0.75\,\mathrm{cm}\) \\
\midrule
\(N_k\) & 15 \\
\midrule
\(\tau_{\mathrm{rec}}\) & \(30\%\) \\
\midrule
\(\gamma\) & \(20\,\mathrm{cm}\) \\
\midrule
\(N_{\mathrm{pos}}\) & 32 \\
\midrule
\(\rho\) & \(10\,\mathrm{cm}\) \\
\midrule
\(d\) & \(10\,\mathrm{cm}\) \\
\bottomrule
\end{tabular}\\[3mm]
}
\textbf{(b)}
\end{minipage}
\vspace{1mm}
\caption{\rev{(a) Experimental results with the graphs of reconstruction percentage achieved by the NBV algorithms. (b) Parameters used for the experiments. Objects have increasing reflectivity from Object A to D.}}
\label{fig:results_total}
\vspace{-5mm}
\end{figure*}

\subsection{Reflective Surface Restoration}
\label{sec:reflective_restoration}
Once a set of reflective regions has been identified, the
recovery process works as follows: (i) generation of candi-
date sensor poses, (ii) definition of an acquisition order based
on a utility function, and (iii) iterative execution until a sufficient level of reconstruction is achieved.

\rev{For the generation of recovery sensor poses, without prior knowledge of the target object, the incidence angle is varied to increase the likelihood of a reliable acquisition. Accordingly, four recovery poses with different incidence angles relative to the detected reflective region are sampled. This strategy is implemented by using the OBB of the detected reflective region to define a local surface frame for pose sampling.}
The OBB $\mathcal{B}_r^\ast \in \mathcal{B}_r$ with the largest volume is selected.
Let $\mathbf{c}_r \in \mathbb{R}^3$ denote the center of the largest face of the selected bounding box $\mathcal{B}_r^\ast$, and let $\mathbf{n}_r \in \mathbb{R}^3$ be the corresponding unit normal vector. We define an orthonormal basis on the tangent plane of the surface by selecting two unit vectors
$\mathbf{t}_1, \mathbf{t}_2 \in \mathbb{R}^3$ that:
\begin{equation}
\mathbf{t}_1^\top \mathbf{n}_r = 0,\quad
\mathbf{t}_2^\top \mathbf{n}_r = 0,\quad
\mathbf{t}_1^\top \mathbf{t}_2 = 0,\quad
\mathbf{t}_2 = \mathbf{n}_r \times \mathbf{t}_1.
\end{equation}
Then, four recovery camera positions $\mathcal{X}_r = \{{x}_r^{i} \}_{i=1}^{4}$
are sampled according to a diamond-shaped layout
in the tangent plane, each one with a center ${o}_k\in \mathbb{R}^3$ defined as: 
\begin{equation}
\mathbf{o}_k = \mathbf{c}_r + d\,\mathbf{n}_r + \rho\,
\left(
\cos\theta_k\,\mathbf{t}_1 + \sin\theta_k\,\mathbf{t}_2
\right),
\label{eq:diamond_positions}
\end{equation}
with $\theta_k \in \left\{0,\frac{\pi}{2},\pi,\frac{3\pi}{2}\right\}$
and where $d>0$ controls the stand-off distance along the normal direction while
$\rho>0$ determines the lateral offset around the surface.
The orientation of each recovery pose is defined by specifying the camera viewing
direction $\mathbf{z}_k$ as:
\begin{equation}
\mathbf{z}_k = \frac{\mathbf{c}_r - \mathbf{o}_k}{\|\mathbf{c}_r - \mathbf{o}_k\|},
\label{eq:lookat_direction}
\end{equation}
so that each pose is oriented toward the target center $\mathbf{c}_r$, resulting in
different angles of incidence with respect to the reflective surface. \rev{The choice of four candidate poses in a diamond-shaped layout is motivated by the need to provide angular diversity by acquiring the target region from sufficiently different positions and orientations, thereby increasing the probability of a successful depth acquisition. Other configurations providing comparable angular diversity could also be effective.} As shown in Fig.~\ref{fig:complete_pipeline}f, for each candidate recovery pose in $\mathcal{X}_r$, the scene is projected back into the image plane following the projection model introduced in Sec.~\ref{sec:view_projection}. In this case, in addition to the ellipsoidal representation of occupied $\mathcal{E}_o$ and frontier regions $\mathcal{E}_f$, the OBB $\mathcal{B}_r^\ast$ associated with the selected reflective surface is also projected. Its projection is obtained by projecting its vertices onto the image plane and computing the corresponding convex hull, yielding a polygonal approximation of area $\mathcal{A}_r$.

For the definition of acquisition order, starting from Eq.~\eqref{eq:view_utility}, an additional term is introduced explicitly biasing the selection toward sensor poses that maximize the visibility of
the reflective surface, while still accounting for occlusions and already observed regions: 
\begin{equation}
U_r(x) =
\sum_{\mathcal{E}_f} w_j\,A_j
-
\sum_{\mathcal{E}_o} w_j\,A_j
+
w_j\,A_r
\label{eq:recovery_view_utility}
\end{equation}
The recovery pose ${x}_r^\ast \in \mathcal{X}_r$ is selected as the candidate that maximizes the utility function $U_r(x)$. 

After the acquisition, we iterate over the remaining recovery poses until a sufficient level of restoration is achieved. To regulate this process, a reconstruction threshold is introduced to achieve a trade-off between reconstruction coverage and acquisition efficiency, thus avoiding an excessive number of recovery views for a single reflective region.
Starting from the selected reflective region $\mathcal{B}_r^\ast$, an enlarged OBB $\mathcal{B}_r^{e\ast}$ is constructed by expanding $\mathcal{B}_r^\ast$ along each axis by twice the resolution of the volumetric discretization employed. The number of voxels contained in the volume of $\mathcal{B}_r^{e\ast}$ defines the maximum capacity $N_{\text{cap}}$. With each new acquisition, the number of newly occupied voxels added inside $\mathcal{B}_r^{e\ast}$ with respect to the initial missing data is computed and denoted as $N_{\text{added}}$.
Inspired by~\cite{reflection_2}, we introduce the recovery index as:
\begin{equation}
\Gamma = \frac{N_{\text{added}}}{N_{\text{cap}}},
\end{equation}
which measures the degree of information recovery as a function of the reconquered volume. \rev{The recovery process is considered successfully concluded when the restoration index exceeds the restoration threshold $\tau_{\text{rec}}$.}


\section{Experiments}

The proposed FRA-NBV method was evaluated in comparison with two reference NBV strategies:
\begin{itemize}
    \item \textbf{Ray-cast Entropy-based method (RE)}~\cite{uf_entropy_2}: A classical approach that selects sensor poses by maximizing volumetric information gain through entropy reduction.
    \item \textbf{Projection-Based method (PB)}~\cite{paper_0}: A faster strategy that
    avoids ray-casting by approximating the object geometry with a compact set of
    ellipsoids and evaluating candidate views through image-plane projections.
\end{itemize}

Fig.~\ref{fig:results_total}a shows the four test objects used in the experiments. In particular, \textbf{Object A} is a non-reflective steel cuboid with simple geometry; \textbf{Object B} is a pneumatic valve featuring mixed materials and moderate geometric complexity, with limited reflective surfaces; \textbf{Object C} is a metallic object with inclined faces and medium reflectivity; \textbf{Object D} has multiple highly reflective steel fittings.

For all methods, the reconstruction process terminates either after $N_{\max}$ iterations or upon detection of stagnation, defined as a relative utility variation below $\tau_{\mathrm{stag}}$ over consecutive acquisitions. 
\rev{The experimental parameters relevant to the evaluation are summarized in Fig.~\ref{fig:results_total}b.}
It is worth noting that, for the proposed FRA-NBV method, each NBV iteration includes the execution of the recovery procedure whenever it is activated. Therefore, recovery poses are not counted as additional iterations, but are integrated within the same NBV
iteration. Under the experimental protocol described above, reconstruction performance was
evaluated using the following metrics: (1) Recovery Activation Count  (Sec.~\ref{sec:Recovery_count}),
(2) Final Reconstruction Coverage (Sec.~\ref{sec:Reconstruction_coverage}), (3) Cumulative and Relative Recovery Contribution (Sec.~\ref{sec:Reconstruction_coverage}), and
(4) Average Iteration Time (Sec.~\ref{sec:average_time}). The definition of each metric and the
corresponding experimental results are presented and discussed in the respective sections.

\subsection{Experimental Setup}

\begin{figure}[!t]
\centering
\begin{minipage}{0.48\columnwidth}
    \centering
    \includegraphics[width=\linewidth]{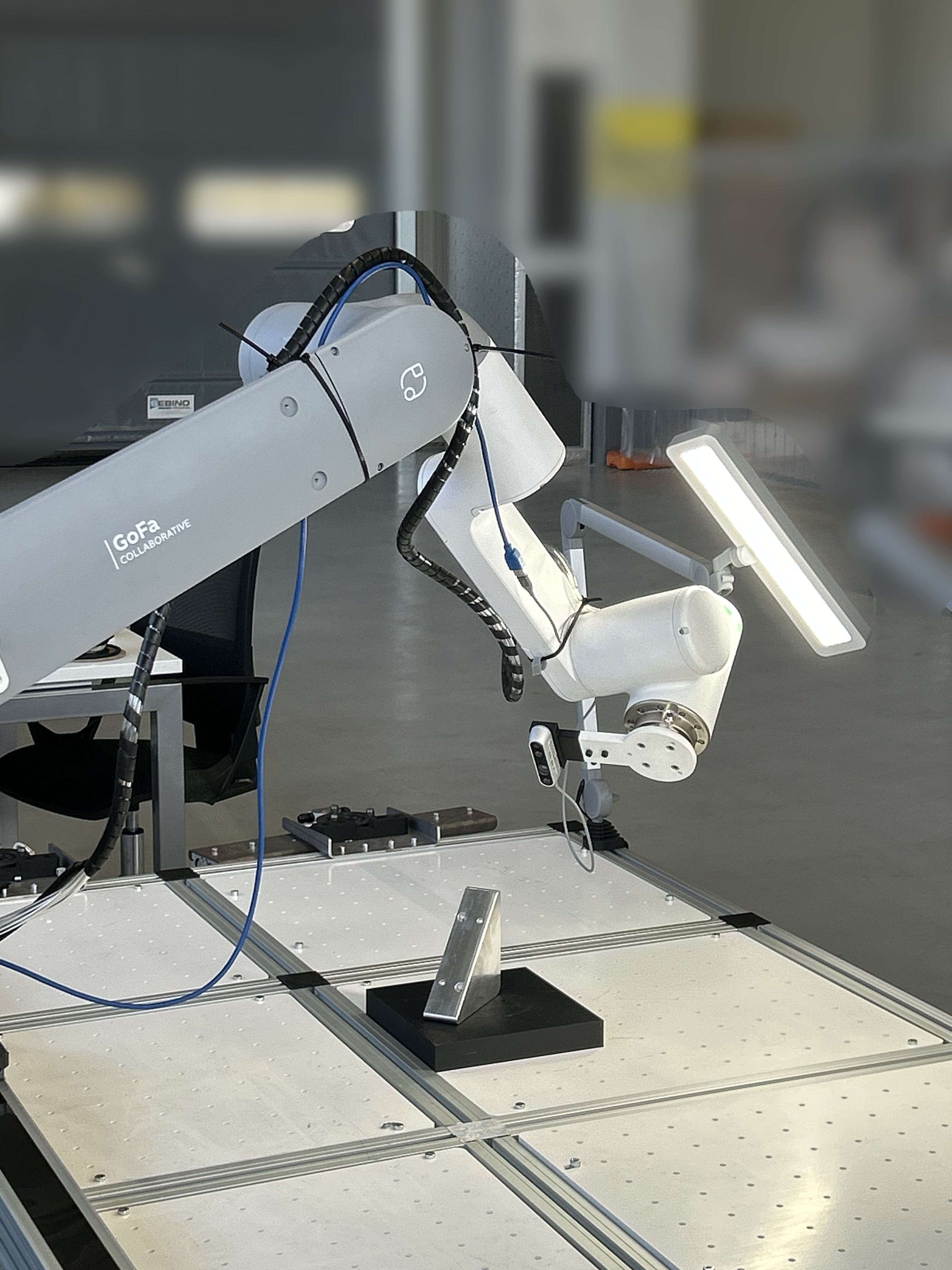}
\end{minipage}
\hfill
\begin{minipage}{0.48\columnwidth}
    \centering
    \includegraphics[width=\linewidth]{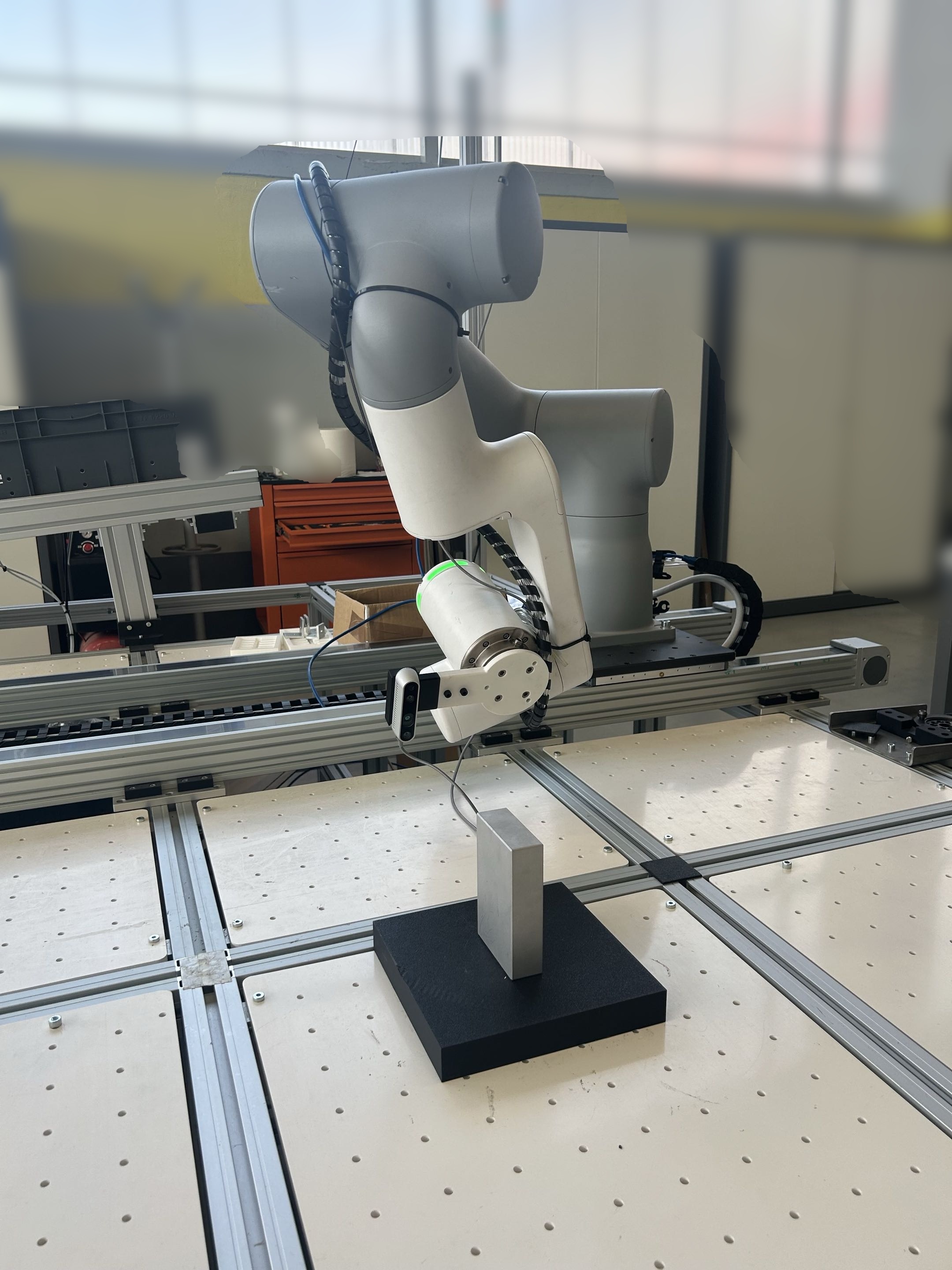}
\end{minipage}
\caption{Experimental setup.}
\vspace{-3mm}
\label{fig:experimental_setup}
\end{figure}

Experiments involved an ABB GoFa 12 manipulator equipped with an Intel
RealSense D435i depth camera mounted on the end-effector. Although the sensor provides
high-resolution depth and RGB data, all RGB streams were disabled
to evaluate the proposed framework under limited perceptual conditions simulating a low-resolution hardware. 
In particular, depth acquisition
was deliberately degraded by applying the maximum decimation factor and additional
cropping, resulting in a final low resolution depth sensor of $60 \times 40$ pixels. An example of the reconstruction process for one of the objects, along with the full experimental setup, is shown in Fig.~\ref{fig:experimental_setup}. \rev{Robot motions were planned with MoveIt! to ensure collision-free trajectories with respect to the robot, the scene geometry, and the online object-estimation boundaries.}
Since ambient illumination significantly affects depth measurements on reflective
surfaces, controlled lighting conditions were adopted in all experiments. A set of fixed
lamps was used to provide consistent illumination of the object workspace, ensuring
repeatability across trials.

\subsection{Recovery Activation Count}
\label{sec:Recovery_count}
\begin{figure}[!t]
\centering
\includegraphics[width=0.7\columnwidth]{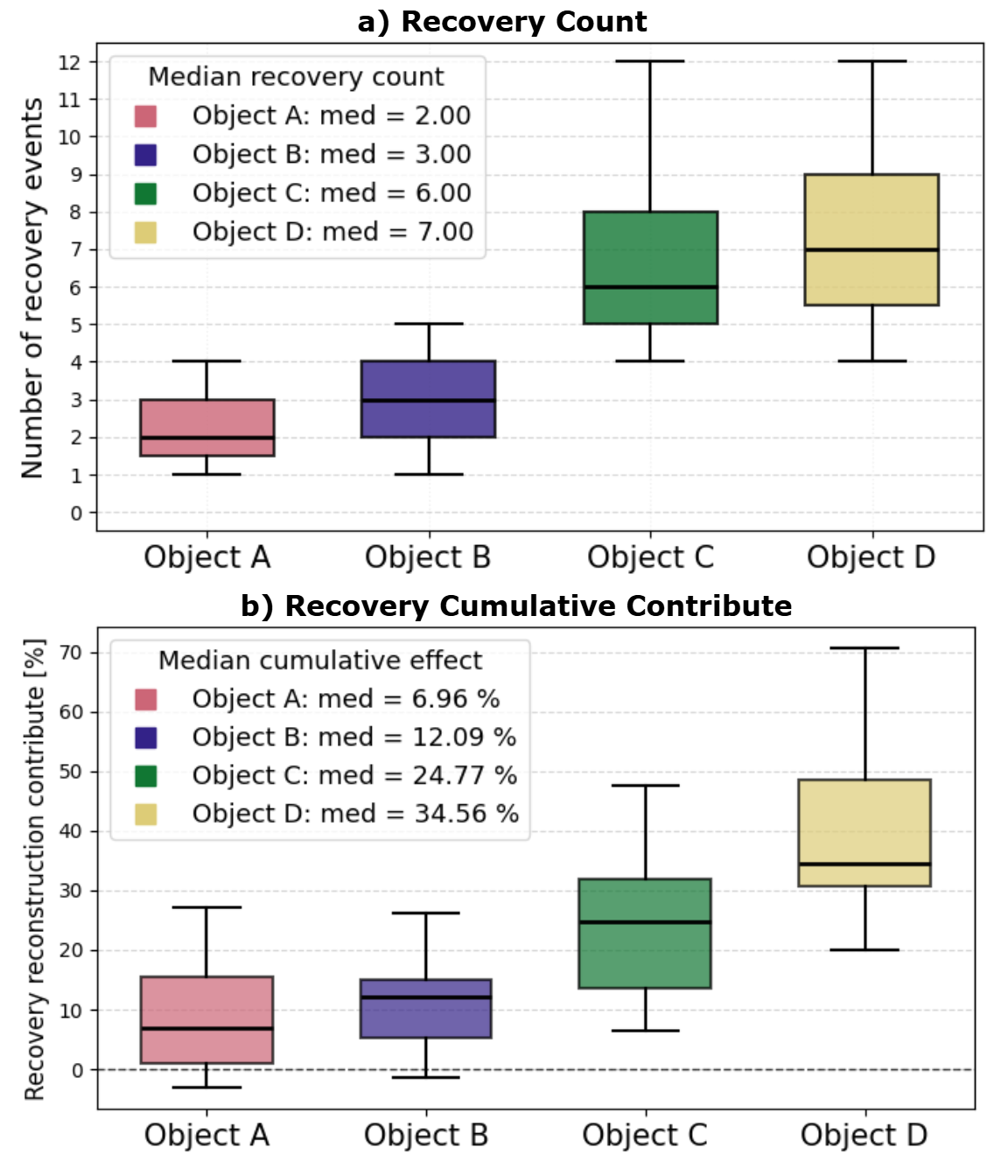}
\caption{Experimental results (over 15 runs): (a) Number of recovery activations occurring during a complete reconstruction. (b) Reconstruction increment (\%) provided by the recovery poses over the full reconstruction.}
\vspace{-5mm}
\label{fig:recovery_count_and_perc}
\end{figure}

This section reports the average number of recovery activations per reconstruction for
FRA-NBV (Fig.~\ref{fig:recovery_count_and_perc}a). The goal is to assess whether the recovery mechanism is selectively triggered by surface reflectivity, rather than being
activated indiscriminately due to algorithmic artifacts.

The results exhibit an increasing trend in the number of recovery activations as
both the reflectivity and geometric complexity of the objects increase. For
\textbf{Object A} and \textbf{Object B}, which are characterized by predominantly
non-reflective surfaces, the average number of activations remains low, with only a few recoveries triggered per reconstruction. This behavior indicates a limited presence of false positives.
In these cases, recovery activations do not originate from reflective behavior; instead, they are mainly attributable to geometric and algorithmic effects. In particular, the ellipsoidal approximation may slightly overestimate the object extent, leading to the classification of boundary-adjacent regions as invalid. This effect is more evident for \textbf{Object A}, while for \textbf{Object B} occasional activations can also be attributed to minor material inhomogeneities that introduce localized reflective effects. Conversely, for the more reflective \textbf{Object C} and \textbf{Object D}, the recovery activation count increases significantly. This confirms that the proposed method is able to correctly identify and target regions affected by unreliable depth measurements caused by surface reflectivity.

\subsection{Reconstruction Coverage and Recovery Contribution}
\label{sec:Reconstruction_coverage}

\begin{figure}[!t]
\centering
\includegraphics[width=0.9\columnwidth]{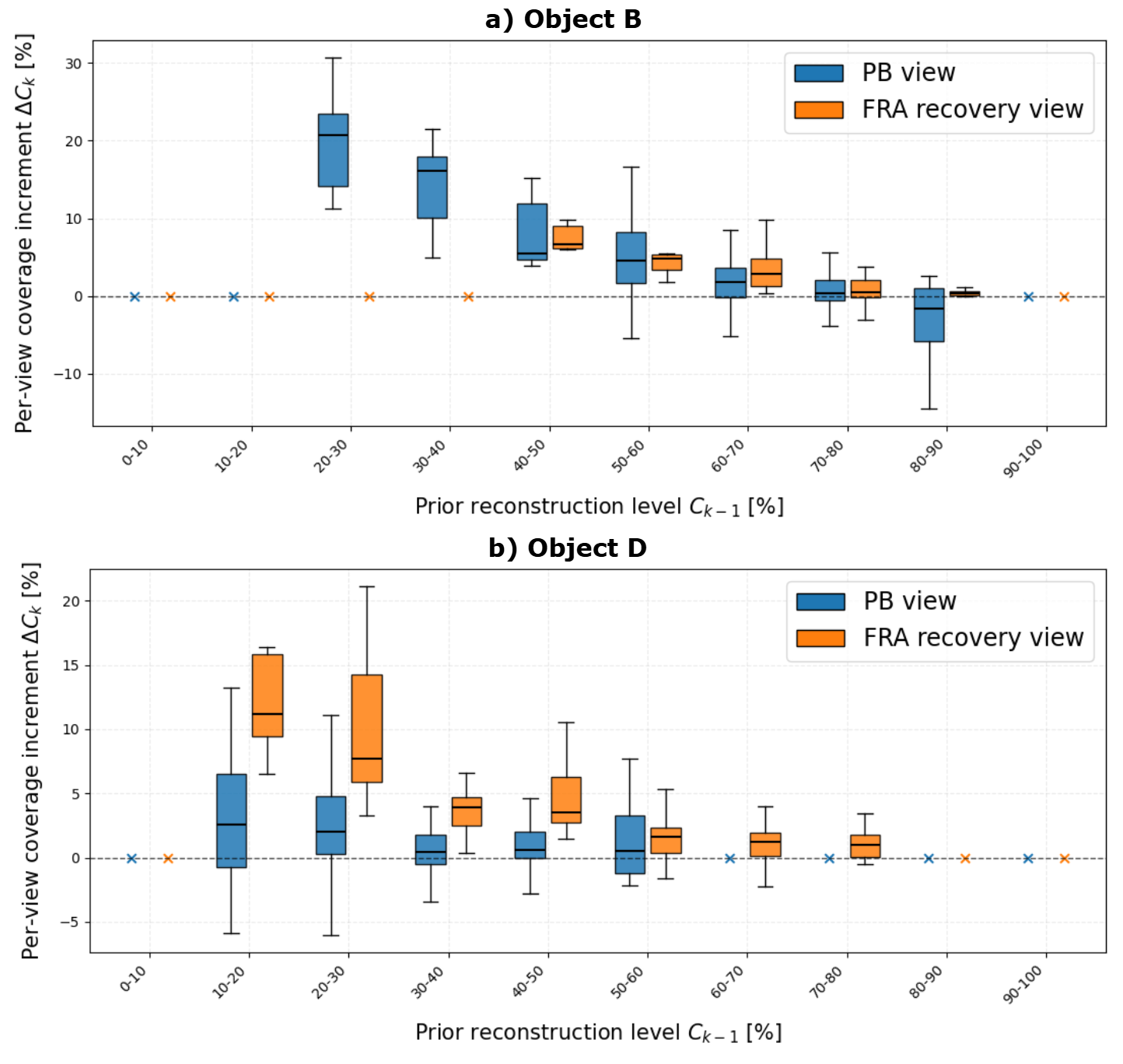}
\caption{Per-view coverage increment $\Delta C_k = C_k - C_{k-1}$ as a function of the prior reconstruction level $C_{k-1}$. 
Boxplots compare standard PB views (blue) and FRA recovery views (orange). 
(a) Non-reflective object: similar contributions. 
(b) Reflective object: higher gain from recovery views. 
``X'' marks indicate insufficient samples.}
\vspace{-5mm}
\label{fig:recovery_view_contribution}
\end{figure}

We evaluate the final reconstruction percentage defined as the
portion of the object surface reconstructed at the end of the process by comparing the final point cloud with the reference CAD model. 
Results are in Fig.~\ref{fig:results_total}a: for \textbf{Object A} and
\textbf{Object B}, which exhibit limited or no reflective properties, all methods achieve comparable reconstruction coverage, with FRA-NBV showing a slightly higher final percentage. 
This indicates that, in the absence of strong reflectivity effects, the
proposed method remains competitive with classical NBV approaches.
\textbf{Object C} shows a clear performance gap between FRA-NBV and the two baseline methods, consistently with the object's moderate reflectivity.
This gap becomes more evident for \textbf{Object D}, the most challenging scenario, where FRA-NBV outperforms the PB method by up to 31\% and the RE method by approximately 56\%. 
These results highlight the robustness of the proposed approach when dealing with highly reflective and geometrically complex objects.
To better interpret these improvements, the cumulative recovery coverage gain was analyzed for the FRA-NBV method. 
This metric measures the increase in surface coverage provided by the recovery poses over the complete reconstruction. 
Results are shown in Fig.~\ref{fig:recovery_count_and_perc}b. For \textbf{Object A} and \textbf{Object B}, recovery poses contribute around 10\% of the final coverage. 
Conversely, for \textbf{Object C} and \textbf{Object D}, the contribution increases up to approximately 30\%. 
In all cases, the additional coverage introduced by recovery
iterations is consistent with the improvements observed in the final reconstruction percentage. These results confirm that the higher reconstruction coverage achieved by FRA-NBV is directly associated with the introduction of recovery poses. 
However, recovery iterations require additional acquisitions, meaning that it is necessary to assess whether the reconstruction improvement owe to a higher informative contribution of FRA-NBV targeting reflective regions.
To address this point, we quantify the relative impact of each acquisition by measuring the percentage coverage increment produced by standard PB views and FRA recovery poses given the same initial reconstruction level.
Let $C_k$ denote the cumulative reconstruction percentage after the $k$th acquisition and define the per-view coverage increment as $\Delta C_k = C_k - C_{k-1}$.
Fig.~\ref{fig:recovery_view_contribution} shows $\Delta C_k$ over the reconstruction coverage achieved prior to that view, $C_{k-1}$. For \textbf{Object B} (negligible reflectivity), the coverage increment provided by recovery acquisitions is comparable to that obtained from PB acquisitions across the entire reconstruction range. 
In contrast, for \textbf{Object D} (the most reflective), recovery acquisitions exhibit a statistically higher coverage increment than standard PB sensor poses. This suggests that the performance gap observed for highly reflective objects cannot be attributed merely to additional acquisitions, but rather to the recovery of otherwise unobservable surface regions.

\subsection{Average Iteration Time}
\label{sec:average_time}

We measure the average time required to complete a single NBV iteration, including view selection, robot motion, and spatial representation update. For the FRA-NBV method, the iteration time includes the execution of the recovery sequence when triggered, from recovery view selection to the acquisition of the final recovery sensor pose. 

\begin{table}[!t]
\centering
\caption{Experimental Results: median and IQR of reconstruction time per acquisition [s]}
\label{tab:time_statistics}
\small
\setlength{\tabcolsep}{5pt}
\renewcommand{\arraystretch}{1.0}
\begin{tabular}{lccc}
\toprule
\textbf{Object} & \textbf{FRA} & \textbf{PB} & \textbf{RE} \\
\midrule
\raisebox{1.2ex}{Object A} & \shortstack{23.09 \\ {[18.84--25.79]}} & \shortstack{\textbf{20.45} \\ \textbf{[18.81--26.11]}} & \shortstack{47.48 \\ {[45.65--48.70]}} \\
\raisebox{1.2ex}{Object B} & \shortstack{25.28 \\ {[22.40--28.19]}} & \shortstack{\textbf{21.72} \\ \textbf{[20.78--24.92]}} & \shortstack{51.76 \\ {[47.50--54.71]}} \\
\raisebox{1.2ex}{Object C} & \shortstack{31.87 \\ {[29.40--38.01]}} & \shortstack{\textbf{21.62} \\ \textbf{[20.64--22.52]}} & \shortstack{46.76 \\ {[45.66--48.01]}} \\
\raisebox{1.2ex}{Object D} & \shortstack{33.10 \\ {[29.01--34.53]}} & \shortstack{\textbf{16.51} \\ \textbf{[13.97--18.47]}} & \shortstack{46.66 \\ {[45.77--49.10]}} \\
\bottomrule
\end{tabular}
\vspace{-4mm}
\end{table}

Table~\ref{tab:time_statistics} shows that FRA-NBV has an average iteration time comparable to PB for \textbf{Object A} and \textbf{Object B}, which are characterized by limited or no reflectivity.
As reflectivity increases, FRA-NBV's iteration time grows because of the larger number of recovery activations. 
Nonetheless, a performance gap remains between projection-based approaches (PB and FRA-NBV) and the RE method. 
The lower iteration times achieved by projection-based strategies confirm that projection-based view evaluation represents a solution for reducing the computational burden of NBV algorithm.

\section{CONCLUSIONS}

We presented FRA-NBV for autonomous 3D reconstruction in the presence of reflective surfaces. 
In non-reflective scenarios, FRA-NBV achieves a reconstruction coverage comparable to standard NBV methods, while outperforming classical raycast-based exploration methods in terms of execution time. 
In highly reflective cases, FRA-NBV achieves up to 31\% higher surface coverage compared to the PB baseline and up to 56\% higher coverage with respect to classical raycast-based exploration strategies. 
Additional qualitative analyses were also conducted on concave geometries involving mutual reflections, showing that FRA-NBV remains less affected by these challenging configurations than the PB-NBV baseline; these results are available on the paper website at \url{https://merlinlaboratory.github.io/FRA-NBV/}.
Future work will explore multimodal extensions integrating additional sensing modalities to further reduce reconstruction failures and support robust active perception.

\bibliographystyle{IEEEtran}
\bibliography{references}

\end{document}